\documentclass{article} % For LaTeX2e
\usepackage{iclr2026_conference,times}

\usepackage{amsmath,amsfonts,bm}

\def\eqref#1{equation~\ref{#1}}
\def\1{\bm{1}}

\DeclareMathAlphabet{\mathsfit}{\encodingdefault}{\sfdefault}{m}{sl}
\SetMathAlphabet{\mathsfit}{bold}{\encodingdefault}{\sfdefault}{bx}{n}

\usepackage{hyperref}
\usepackage{url}

\usepackage[T1]{fontenc}
\usepackage{lmodern}
\usepackage{microtype}
\usepackage{amsmath,amssymb}
\usepackage{booktabs}
\usepackage{graphicx}
\usepackage{xcolor}
\usepackage{tikz}
\usetikzlibrary{arrows.meta,calc,fit,positioning}
\usepackage{enumitem}
\usepackage[font=small,labelfont=bf]{caption}
\usepackage{subcaption}
\providecommand{\phantomcaption}{\refstepcounter{subfigure}}
\usepackage{float}

\graphicspath{{figures/}}
\newenvironment{promptbox}[1]
  {\par\medskip\noindent{\footnotesize\bfseries #1}\par\nobreak\smallskip
   \hrule\vspace{0.3em}\small}
  {\par\vspace{0.15em}\hrule\par\medskip}

\definecolor{domainblue}{HTML}{1D4E89}
\definecolor{framingorange}{HTML}{9A4D00}
\definecolor{utilitygreen}{HTML}{246A39}

\title{Training Language Models to Be Coherent \\ Decision-Makers}
\iclrfinalcopy
\iclrfinalcopy
\author{
Khurram Yamin \\
Carnegie Mellon University \\
\texttt{kyamin@andrew.cmu.edu}
\And
Xavier Fernandes \\
Microsoft \\
\texttt{xfernandes@microsoft.com}
\And
Paul Koch \\
Microsoft \\
\texttt{paulkoch@microsoft.com}
\AND
Bryan Wilder \\
Carnegie Mellon University \\
\texttt{bwilder@andrew.cmu.edu}
\And
Eric Horvitz \\
Microsoft \\
\texttt{horivitz@microsoft.com}
}

\begin{document}

\maketitle

\begin{abstract}
Reliable decision-making requires more than accurate prediction: a model must preserve its beliefs, apply the relevant utilities, and recognize when the information needed to justify an action is missing. We study whether language models can learn this decision procedure from supervised fine-tuning and generalize it across domains and differing natural-language expressions of the decision challenge. Across 20 datasets, we explore challenges of belief instability and decision-making errors by first eliciting probabilities of  outcomes and then varying only the utilities and the framing of the decision problems, while holding the evidence fixed. We train models to preserve elicited beliefs while selecting the action that maximizes expected utility, and evaluate transfer to unseen application domains, held-out framings, and different classes of payoff structures. We further introduce incomplete-information settings in which required utilities are withheld and replaced with irrelevant text, testing whether models can distinguish missing decision-relevant information from merely additional context. We find that targeted fine-tuning substantially improves coherent decision-making and that in many situations, learning transfers across domains and framings to situations unobserved during training. Further, models trained for decidability learn to identify when  action cannot be justified based on missing information. Finally, we show the value of a routed system that considers separately the recognition of decision completeness and utility-sensitive decision execution.
\end{abstract}

\section{Introduction}
\label{sec:introduction}

Accurate prediction is not sufficient for good decision-making. A language
model may assess the likelihood of an outcome well yet still choose poorly if
the stakes distort its belief, if it misapplies the consequences of its actions,
or if it fails to notice that information needed to justify an action is
missing. Decision theory separates these steps: beliefs about uncertain
outcomes and utilities over their consequences combine into an action
\citep{Savage1972Foundations, VonNeumann1944-VONTOG-4}. A clinical assistant
must weigh the risk of a complication against the cost of escalating care, and
a fraud system the likelihood of abuse against the cost of blocking a
legitimate transaction. The same risk justifies different actions under
different utilities, and no action is justified when the utilities are not
stated. Because evidence and utilities arrive in varying natural language, a
model must also recover the same decision problem from different formulations
and ignore details that do not affect the choice.

Work on uncertainty in language models has mostly studied calibration in
isolation from the decisions it supports
\citep{kadavath2022languagemodelsmostlyknow,lin2022teachingmodelsexpressuncertainty,cruz2024evaluatinglanguagemodelsrisk}.
Evaluations that test stated beliefs against actions find that frontier models
are imperfect but fairly coherent decision-makers, although their beliefs and
implied preferences can still diverge from their behavior
\citep{yamin2026agentssaythinganother,pal2026knowingknowenoughlarge,yamin2026revealedpreferencesclarifyllm,mazeika2025utilityengineeringanalyzingcontrolling}.
Many decision-support settings, however, cannot use a frontier model: the data must stay
on premises, each call is bound by cost and latency, or the system must run on
local hardware. We therefore ask: \emph{can small language
models be trained to follow a coherent decision procedure, and does what they
learn generalize beyond the framings and domains seen during training?}

We deliberately start with a simple setting: binary 
decisions with asymmetric error costs, where the optimal action is a threshold
on the belief set by the two costs, so, given a stated belief, exactly one
action is correct. We would expect current small models to handle this well, yet they 
fail starkly, acting against their own stated beliefs and shifting those
beliefs once payoffs appear. Across 20 tabular
domains rendered as prose, we first elicit each model's belief from the evidence alone, then hold
the evidence fixed and vary only the utilities and the natural-language framing
of the decision. This separates two failures that final-answer accuracy
conflates: changing one's belief when only the stakes change, and choosing an
action that does not follow from one's stated belief. We then train with
supervised fine-tuning (SFT)
\citep{ouyang2022traininglanguagemodelsfollow,wei2022finetunedlanguagemodelszeroshot}.
The target for each decision prompt is the model's own independently elicited
belief together with the expected-utility-maximizing action under that belief,
written either as a short structured answer (Direct) or as a chain of thought
that carries the belief through the payoff calculation (CoT), so supervision
teaches the mapping from belief and consequences to action without imposing
external probability labels.

Because strong performance on familiar formats need not mean the decision-making procedure has
been learned
\citep{kang2024learningdynamicsrevealgeneralization,ren2025learningdynamicsllmfinetuning},
we hold out entire application domains, individual framings, and whole classes
of payoff structure from training. Targeted SFT makes small models
substantially more coherent, with gains that transfer to unseen domains and to
substantively different descriptions of the same utilities. Finally, we ask whether models can recognize when a decision is underspecified.
We withhold one or both error costs and replace them with irrelevant text of
matched length, so a coherent model must report its belief, name the missing
information, and decline to act \citep{li2025questbenchllmsaskright}. Models
fine-tuned for this \emph{decision decidability} learn to identify what is
missing but decide less well on the complete cases, so we compose two
specialists into a \emph{routed system}: a decidability gate that reports the
belief and either names the missing cost or forwards complete cases to a
complete-information decision model. The routed system gives the best
end-to-end accuracy for five of six models with almost no loss in belief
stability.

\section{Related Work}
\label{sec:related-work}

\paragraph{Beliefs, utilities, and decisions.}
Work on language-model uncertainty asks whether probabilities or confidence
reports are calibrated and predictive of correctness
\citep{lin2022teachingmodelsexpressuncertainty,kadavath2022languagemodelsmostlyknow,
tian2023justaskcalibrationstrategies,xiong2024llmsexpressuncertaintyempirical,
cruz2024evaluatinglanguagemodelsrisk}, and complementary work tests whether
elicited beliefs agree with downstream behavior
\citep{freedman2025exploringpotentiallargelanguage,
yamin2026agentssaythinganother,pal2026knowingknowenoughlarge}. Behavioral
evaluations reveal biases across lotteries, cognitive tasks, and framings
\citep{binz2022usingcognitivepsychologyunderstand,
echterhoff2024cognitivebiasdecisionmakingllms}, and related work elicits model
preference systems \citep{mazeika2025utilityengineeringanalyzingcontrolling} or
infers loss functions from observed choices
\citep{yamin2026revealedpreferencesclarifyllm,yamin2026highstakesdecisionslanguagemodels}. Our objective is different:
utilities are explicitly supplied, in varying wording, and models must
implement them faithfully without changing the underlying belief. Unlike
decision-focused learning
\citep{wilder2018meldingdatadecisionspipelinedecisionfocused}, which modifies
predictors for downstream optimization, we preserve the model's belief and
train the mapping from belief and consequences to action.

\paragraph{Learning procedures that generalize.}
Instruction tuning and SFT teach new behaviors from targeted supervision
\citep{ouyang2022traininglanguagemodelsfollow,
wei2022finetunedlanguagemodelszeroshot,zhou2023limaalignment}, and rationale
supervision and chain-of-thought improve multi-step reasoning
\citep{hsieh2023distillingstepbystepoutperforminglarger,
wei2023chainofthoughtpromptingelicitsreasoning}. Better performance after
training need not mean the intended procedure was learned: fine-tuning can
reinforce surface-level associations
\citep{ren2025learningdynamicsllmfinetuning}, and apparent gains may reflect
memorization \citep{kang2024learningdynamicsrevealgeneralization}. Our
holdouts of whole domains, framings, and payoff classes are designed to test
for this.

\paragraph{Incomplete information and specialization.}
Recent work studies whether models recognize missing information, acquire it,
or cope when it is spread across turns
\citep{laban2025llmslostmultiturnconversation,
wu2025collabllmpassiverespondersactive,li2025questbenchllmsaskright}. We
consider a decision-theoretic form of incompleteness: evidence may support a
probability while a missing utility leaves the action undecidable, and
replacing that utility with extraneous text tests whether the model recognizes
decision relevance rather than the apparent completeness of a prompt.
Learning-to-defer methods transfer instances to experts
\citep{mozannar2021consistentestimatorslearningdefer,
wilder2020learningcomplementhumans}, and LLM routers select among models by
expected quality or cost \citep{lu2023routingexpertefficientrewardguided,
chai2024expertworthtokensynergizing,ong2025routellmlearningroutellms}; our
gate instead asks whether the decision is well specified and passes only
complete cases to a specialist, separating decidability recognition from
decision execution.

\section{Methods}

\begin{figure}[t]
  \centering
  \begingroup
  \definecolor{overviewink}{HTML}{263340}
  \definecolor{overviewline}{HTML}{A9B4BF}
  \definecolor{overviewpanel}{HTML}{F6F8FA}
  \definecolor{solverblue}{HTML}{255D8A}
  \definecolor{solverlight}{HTML}{E8F1F8}
  \definecolor{gateorange}{HTML}{A85817}
  \definecolor{gatelight}{HTML}{FBEEDF}
  \definecolor{completegreen}{HTML}{33734A}
  \definecolor{completelight}{HTML}{E9F4EC}
  \definecolor{missingred}{HTML}{A34545}
  \definecolor{missinglight}{HTML}{F9EAEA}
  \definecolor{framingorange}{HTML}{9A4D00}

  \resizebox{\linewidth}{!}{%
  \begin{tikzpicture}[
    x=1cm,y=1cm,
    every node/.style={font=\sffamily\scriptsize,text=overviewink,align=center},
    flow/.style={-{Latex[length=2.1mm,width=1.4mm]},line width=0.75pt,draw=overviewink},
    softflow/.style={-{Latex[length=1.9mm,width=1.25mm]},line width=0.65pt,draw=overviewline},
    box/.style={rounded corners=2.2pt,draw=overviewline,line width=0.65pt,
      fill=white,inner sep=4.5pt,minimum height=1.10cm},
    stage/.style={rounded corners=3pt,line width=0.8pt,inner sep=5pt,
      minimum height=1.25cm},
    badge/.style={rounded corners=5pt,inner xsep=6pt,inner ysep=2.7pt,
      font=\sffamily\scriptsize\bfseries},
    slot/.style={rounded corners=1.7pt,line width=0.55pt,
      minimum width=0.95cm,minimum height=0.34cm,
      inner sep=1.1pt,font=\sffamily\tiny},
    known/.style={slot,draw=completegreen!60,fill=completelight},
    missing/.style={slot,draw=missingred!70,fill=missinglight,
      text=missingred,font=\sffamily\tiny\bfseries},
    scribble/.style={draw=overviewink!28,line width=0.42pt},
    tinylab/.style={font=\sffamily\tiny,text=overviewink!78}
  ]

    % ============================================================
    % OUTER PANELS
    % ============================================================

    \filldraw[
      fill=overviewpanel,
      draw=overviewline,
      line width=0.55pt,
      rounded corners=4pt
    ] (0,5.90) rectangle (17.05,9.50);

    \filldraw[
      fill=overviewpanel,
      draw=overviewline,
      line width=0.55pt,
      rounded corners=4pt
    ] (0,0) rectangle (17.05,5.65);

    % ============================================================
    % PANEL (a): COMPLETE-INFORMATION DECISION LEARNING
    % ============================================================

    \node[
      anchor=west,
      font=\sffamily\small\bfseries
    ] at (0.28,9.18)
      {(a) Complete-information decision learning};

    \node[box,text width=1.75cm] (evidence) at (1.30,7.55)
      {\textbf{Case evidence}\\[-1pt]
       Covariates $x$\\
       binary event $E$};

    \node[
      box,
      text width=1.80cm,
      fill=completelight,
      draw=completegreen!55
    ] (utilities) at (4.10,7.55)
      {\textbf{Complete utilities}\\[-1pt]
       $c_{\mathrm{FP}}$ and\\
       $c_{\mathrm{FN}}$ given};

    \node[
      box,
      text width=2.15cm,
      fill=framingorange!7,
      draw=framingorange!60
    ] (framings) at (7.05,7.55)
      {\textbf{8 equivalent natural language\\ framings}};

    \node[
      stage,
      text width=3.15cm,
      fill=solverblue,
      draw=solverblue,
      text=white
    ] (solverA) at (11.05,7.55)
      {\textbf{Complete-information solver}\\[-1pt]
       preserve belief; recover $p^\star$;\\
       choose \textsc{Act} or \textsc{Hold}};

    \node[
      box,
      text width=2.35cm,
      fill=solverlight,
      draw=solverblue!55
    ] (decisionA) at (15.05,7.55)
      {\textbf{Decision output}\\[-1pt]
       belief $p'$ and utility-\\
       maximizing action};

    \draw[flow] (evidence.east) -- (utilities.west);
    \draw[flow] (utilities.east) -- (framings.west);
    \draw[flow] (framings.east) -- (solverA.west);
    \draw[flow] (solverA.east) -- (decisionA.west);

    \node[
      badge,
      fill=solverlight,
      text=solverblue,
      draw=solverblue!45
    ] (solvertrain) at (11.05,8.85)
      {SFT};

    \draw[softflow] (solvertrain) -- (solverA);

    % ============================================================
    % PANEL (b): DECISION DECIDABILITY
    % ============================================================

    \node[
      anchor=west,
      font=\sffamily\small\bfseries
    ] at (0.28,5.33)
      {(b) Decision decidability and routed specialization};

    % ============================================================
    % LOWER INTERNAL PANELS
    % ============================================================

    \filldraw[
      fill=white,
      draw=overviewline,
      line width=0.45pt,
      rounded corners=3pt
    ] (0.40,0.35) rectangle (7.15,4.88);

    \filldraw[
      fill=white,
      draw=overviewline,
      line width=0.45pt,
      rounded corners=3pt
    ] (7.45,0.35) rectangle (10.55,4.88);

    \filldraw[
      fill=white,
      draw=overviewline,
      line width=0.45pt,
      rounded corners=3pt
    ] (10.85,0.35) rectangle (16.70,4.88);

    % ============================================================
    % INTERNAL HEADINGS
    % ============================================================

    \node[
      anchor=west,
      font=\sffamily\scriptsize\bfseries,
      text=overviewink!90
    ] at (0.62,4.62)
      {Prompt and decision information};

    \node[
      anchor=west,
      font=\sffamily\scriptsize\bfseries,
      text=overviewink!90
    ] at (7.72,4.62)
      {Decidability};

    \node[
      anchor=west,
      font=\sffamily\scriptsize\bfseries,
      text=overviewink!90
    ] at (11.05,4.62)
      {Routed execution};

    % ============================================================
    % DECISION PROMPT
    % ============================================================

    \node[
      box,
      minimum width=2.28cm,
      minimum height=3.30cm,
      anchor=north west
    ] (prompt) at (0.72,4.25) {};

    \node[
      anchor=west,
      font=\sffamily\scriptsize\bfseries
    ] at (0.90,3.99)
      {Decision prompt};

    \node[tinylab,anchor=west] at (0.90,3.74)
      {Case context};

    \draw[scribble] (0.90,3.53) -- (2.24,3.53);
    \draw[scribble] (0.90,3.36) -- (2.40,3.36);
    \draw[scribble] (0.90,3.19) -- (2.14,3.19);
    \draw[scribble] (0.90,3.02) -- (2.34,3.02);

    \node[tinylab,anchor=west] at (0.90,2.74)
      {Question / actions};

    \draw[scribble] (0.90,2.53) -- (2.32,2.53);
    \draw[scribble] (0.90,2.36) -- (2.10,2.36);

    \node[
      rounded corners=1.8pt,
      draw=gateorange!55,
      fill=gatelight,
      minimum width=1.78cm,
      minimum height=0.58cm
    ] at (1.73,1.90) {};

    \node[
      tinylab,
      text=gateorange
    ] at (1.73,1.90)
      {utility / payoff\\information};

    \draw[scribble] (0.90,1.46) -- (2.24,1.46);
    \draw[scribble] (0.90,1.29) -- (2.35,1.29);
    \draw[scribble] (0.90,1.12) -- (2.07,1.12);

    % made horizontal
    \draw[flow] (prompt.east) -- ++(0.38,0);

    % ============================================================
    % REQUIRED DECISION STRUCTURE
    % ============================================================

    \node[
      font=\sffamily\scriptsize\bfseries
    ] at (5.30,4.13)
      {Required decision structure};

    \node[
      box,
      minimum width=1.00cm,
      minimum height=0.58cm
    ] (root) at (5.30,3.62)
      {Decision};

    \coordinate (fork) at (5.30,3.30);
    \coordinate (act)  at (4.72,2.96);
    \coordinate (hold) at (5.88,2.96);

    \draw[
      overviewink,
      line width=0.75pt
    ] (root.south) -- (fork);

    \draw[
      overviewink,
      line width=0.75pt
    ] (4.72,3.30) -- (5.88,3.30);

    \draw[flow] (4.72,3.30) -- (act);
    \draw[flow] (5.88,3.30) -- (hold);

    \filldraw[
      draw=overviewink,
      fill=white,
      line width=0.55pt
    ] (act) circle (0.08);

    \filldraw[
      draw=overviewink,
      fill=white,
      line width=0.55pt
    ] (hold) circle (0.08);

    \node[tinylab,anchor=east] at (4.55,2.96)
      {\textsc{Act}};

    \node[tinylab,anchor=west] at (6.05,2.96)
      {\textsc{Hold}};

    \node[known] (fp) at (4.72,2.46)
      {$c_{\mathrm{FP}}$};

    \node[known] (fn) at (5.88,2.46)
      {$c_{\mathrm{FN}}$};

    \draw[flow] (act) -- (fp.north);
    \draw[flow] (hold) -- (fn.north);

    % ============================================================
    % A--D CONDITIONS
    % ============================================================

    \node[
      tinylab,
      font=\sffamily\tiny\bfseries
    ] at (4.72,2.00)
      {FP cost};

    \node[
      tinylab,
      font=\sffamily\tiny\bfseries
    ] at (5.88,2.00)
      {FN cost};

    \node[
      font=\sffamily\scriptsize\bfseries,
      text=missingred
    ] at (4.05,1.70) {A};

    \node[missing] at (4.72,1.70) {?};
    \node[known] at (5.88,1.70) {$c_{\mathrm{FN}}$};

    \node[
      font=\sffamily\scriptsize\bfseries,
      text=missingred
    ] at (4.05,1.34) {B};

    \node[known] at (4.72,1.34) {$c_{\mathrm{FP}}$};
    \node[missing] at (5.88,1.34) {?};

    \node[
      font=\sffamily\scriptsize\bfseries,
      text=missingred
    ] at (4.05,0.98) {C};

    \node[missing] at (4.72,0.98) {?};
    \node[missing] at (5.88,0.98) {?};

    \node[
      font=\sffamily\scriptsize\bfseries,
      text=completegreen
    ] at (4.05,0.62) {D};

    \node[known] at (4.72,0.62) {$c_{\mathrm{FP}}$};
    \node[known] at (5.88,0.62) {$c_{\mathrm{FN}}$};

    % ============================================================
    % DECIDABILITY MODEL
    % ============================================================

    \draw[flow] (7.12,2.43) -- (7.42,2.43);

    \node[
      badge,
      fill=gatelight,
      text=gateorange,
      draw=gateorange!45
    ] (sftB) at (8.98,3.80)
      {SFT on A--D};

    \node[
      stage,
      text width=2.10cm,
      fill=gateorange,
      draw=gateorange,
      text=white,
      minimum height=1.38cm
    ] (gate) at (8.98,2.43)
      {\textbf{Decidability model}\\[-1pt]
       predict A--D; report $p'$;\\
       act when complete info};

    \draw[softflow] (sftB) -- (gate);

    % ============================================================
    % ROUTED EXECUTION
    % ============================================================

    \node[
      box,
      text width=2.00cm,
      fill=missinglight,
      draw=missingred!65,
      minimum height=0.95cm
    ] (stop) at (12.65,3.55)
      {\textbf{A--C: incomplete tree}};

    \node[
      box,
      text width=2.10cm,
      fill=solverlight,
      draw=solverblue!55,
      minimum height=1.05cm
    ] (generalist) at (15.50,3.55)
      {\textbf{Generalist Model}\\[-1pt]
       use decidability\\
       output directly};

    \node[
      box,
      text width=2.35cm,
      minimum height=1.82cm,
      fill=completelight,
      draw=completegreen!65
    ] (fulltree) at (12.65,1.38) {};

    \node[
      font=\sffamily\scriptsize\bfseries,
      text=completegreen
    ] at (12.65,2.06)
      {D: complete tree};

    \node[
      stage,
      text width=2.10cm,
      fill=solverblue,
      draw=solverblue,
      text=white,
      minimum height=1.35cm
    ] (specialist) at (15.50,1.38)
      {\textbf{Specialist solver}\\[-1pt]
       reuse SFT\\
       model from (a)};

    % ============================================================
    % ROUTING ARROWS
    % ============================================================

    % Clean vertical branch spine, matching the standalone version.
    \coordinate (gatebranch) at (11.00,2.43);

    % Gate enters the branch without an intermediate arrowhead.
    \draw[
      overviewink,
      line width=0.75pt
    ] (gate.east) -- (gatebranch);

    % Vertical split.
    \draw[
      overviewink,
      line width=0.75pt
    ] (11.00,1.38) -- (11.00,3.55);

    % Horizontal branch arrows.
    \draw[flow] (11.00,3.55) -- (stop.west);
    \draw[flow] (11.00,1.38) -- (fulltree.west);

    % Horizontal output arrows.
    \draw[flow] (stop.east) -- (generalist.west);
    \draw[flow] (fulltree.east) -- (specialist.west);

    % Only diagonal connector.
    \draw[flow]
      (fulltree.north east) --
      (generalist.south west);

    % ============================================================
    % MINI COMPLETE TREE
    % ============================================================

    \coordinate (droot)  at (12.65,1.56);
    \coordinate (dsplit) at (12.65,1.30);
    \coordinate (dleft)  at (12.15,1.02);
    \coordinate (dright) at (13.15,1.02);

    \draw[
      overviewink,
      line width=0.55pt
    ] (droot) -- (dsplit);

    \draw[
      overviewink,
      line width=0.55pt
    ] (12.15,1.30) -- (13.15,1.30);

    \draw[
      overviewink,
      line width=0.55pt
    ] (12.15,1.30) -- (dleft);

    \draw[
      overviewink,
      line width=0.55pt
    ] (13.15,1.30) -- (dright);

    \filldraw[
      draw=overviewink,
      fill=white,
      line width=0.45pt
    ] (droot) circle (0.05);

    \filldraw[
      draw=overviewink,
      fill=white,
      line width=0.45pt
    ] (dleft) circle (0.05);

    \filldraw[
      draw=overviewink,
      fill=white,
      line width=0.45pt
    ] (dright) circle (0.05);

    \node[
      known,
      minimum width=0.82cm,
      minimum height=0.31cm
    ] at (12.06,0.63)
      {$c_{\mathrm{FP}}$};

    \node[
      known,
      minimum width=0.82cm,
      minimum height=0.31cm
    ] at (13.24,0.63)
      {$c_{\mathrm{FN}}$};

  \end{tikzpicture}%
  }
  \endgroup

  \caption{Overview of the benchmark and training pipeline. (a) We first
  fine-tune a complete-information solver to preserve its elicited belief and
  apply the utility-implied decision rule across eight equivalent framings of
  each case--utility pair. (b) We then study decision decidability under
  incomplete utility information. The decidability model identifies which
  information condition is present, reports a belief, withholds action for
  incomplete cases, and routes complete decision problems to the specialist
  solver.}
  \label{fig:method-overview}
\end{figure}
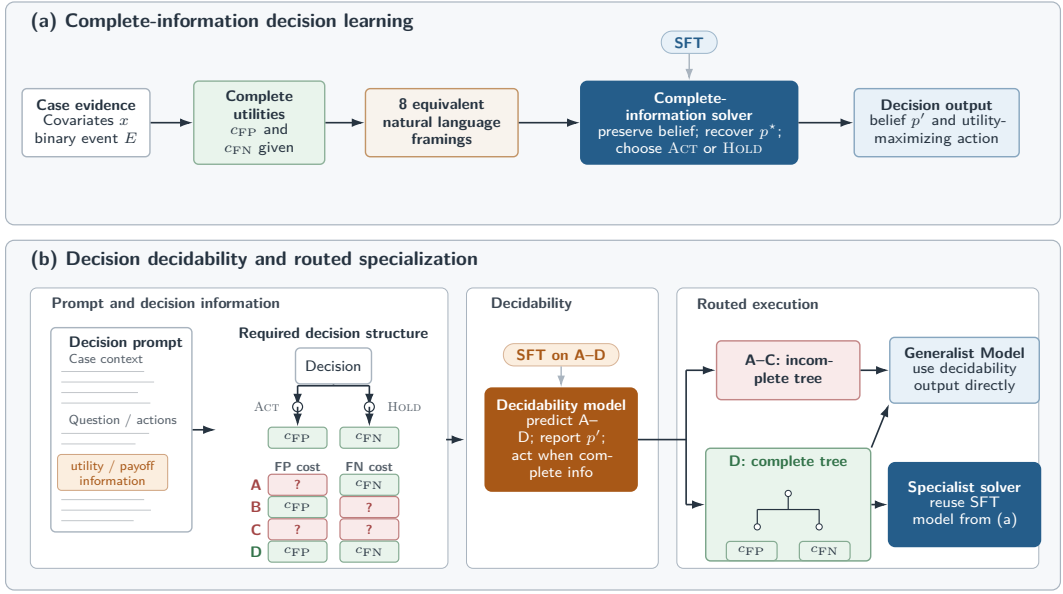

Figure~\ref{fig:method-overview} summarizes our design. We first study
complete-information decisions, where every error is either a shift in the
stated belief or a misapplication of the utilities, and train models with SFT
for this setting. We then withhold utility information, so a model must first
judge whether the decision is decidable and, if not, name what is missing.

\subsection{Decision setting}
\label{sec:decision-setting}

\paragraph{Binary decisions under asymmetric error costs.}
We consider binary decisions with asymmetric error costs, a deliberately simple
setting in which base models nonetheless fail (Section~\ref{sec:results}) and
which covers most decisions in our domains. Each case contains evidence \(x\) about an unobserved binary event \(E\) and
asks the model to choose between \textsc{Act} and \textsc{Hold}, where acting is
right when \(E\) is true. Acting when \(E\) is false costs \(c_{\mathrm{FP}}\)
and holding when \(E\) is true costs \(c_{\mathrm{FN}}\), so under belief
\(p=P(E\mid x)\) the expected costs are $L(\textsc{Act};p)=(1-p)c_{\mathrm{FP}}$
and $L(\textsc{Hold};p)=p\,c_{\mathrm{FN}}$, which are equal at
\begin{equation}
  p^\star=\frac{c_{\mathrm{FP}}}
  {c_{\mathrm{FP}}+c_{\mathrm{FN}}},
  \qquad
  d^\star(p)=
  \begin{cases}
    \textsc{Act}, & p\geq p^\star,\\
    \textsc{Hold}, & p<p^\star.
  \end{cases}
  \label{eq:threshold}
\end{equation}
Acting minimizes expected cost above this threshold and holding below it.
Reward-based presentations reduce to the same form: subtracting each action's
payoff from the best payoff in that state recovers the same costs and threshold.

\paragraph{Stability of beliefs.}
For every case, we first query the model in a context containing \(x\) but no
actions or payoffs. We average five independent samples to obtain the model's
reference belief 
\begin{equation}
  p_E(x)=\frac{1}{5}\sum_{r=1}^{5}\widehat P_r(E\mid x).
  \label{eq:reference-belief}
\end{equation}
We then open a fresh context with the same evidence and a decision problem, in
which the model reports a decision-time belief \(p'\) and an action
\(d_{\mathrm{model}}\) in one parseable final block. The elicitation is never
part of the decision prompt, and reporting \(p'\) with the action lets us test
whether the action follows from the belief stated at decision time.

\subsection{Complete-information benchmark}
\label{sec:complete-benchmark}

\paragraph{Domains and covariates.}
We construct natural-language cases from 20 public tabular binary-prediction
datasets spanning finance, medicine, education, marketing, equipment failure,
and online behavior (Appendix~\ref{app:domains}). A case renders selected
covariates from one row as prose and defines the event from the dataset
outcome; we sample 10 cases per domain, for 200 evidence contexts. Outcome
labels are used to construct and audit cases, never as SFT targets.

\paragraph{Utilities and framings.}
We instantiate five cost pairs, $(c_{\mathrm{FP}},c_{\mathrm{FN}})
\in\{(10,1),(10,2),(1,1),(2,10),(1,10)\}$, which induce
\(p^\star\in\{0.909,0.833,0.500,0.167,0.091\}\) (rounded). Each case--utility
pair is rendered in eight framings from four structural classes:
\emph{protective action} (intervention, insurance: a certain protective cost
against an event-contingent loss), \emph{hedge or offset} (betting, trading: a
stake or contract that offsets exposure to the event), \emph{sure versus
contingent payoff} (contingency, holdout: a certain payment against an
event-contingent one), and \emph{abstract ratio} (loss-ratio, utility-ratio:
relative costs of errors or rewards of correct actions). The action names and
payoff presentation change, but every rendering preserves \(p^\star\). The
threshold itself never appears in a prompt; each framing states costs or
payoffs in its own terms, and we recover \(p^\star\) from every rendered prompt
as a construction check. Crossing the 200 cases with five utilities, eight
framings, and two decision repetitions yields 16,000 decision prompts per
prompting condition; separately, we collect five belief elicitations per case.

\paragraph{Example items.}
Figure~\ref{fig:examples} shows an abridged complete-information item and the
same case with one error cost withheld (Section~\ref{sec:decidability});
Appendix~\ref{app:prompts} gives every template verbatim.

\begin{figure}[t]
  \centering
  \begingroup
  \setlength{\fboxsep}{6pt}
  \fcolorbox{black!25}{black!2}{%
    \begin{minipage}{0.96\linewidth}
      \footnotesize
      \begin{minipage}[t]{0.48\linewidth}
        \textbf{Complete information}\par\smallskip
        {\color{domainblue}\textbf{Domain/covariates.} A 27-year-old student
        entered with a rank of 120/200, does not receive a scholarship, and
        passed 0/4 first-semester courses and 0/5 second-semester courses.\par}\smallskip
        {\color{framingorange}\textbf{Framing.} What is the probability that
        the student drops out? Choose \textsc{Intervene} with academic support
        or \textsc{Hold}.\par}\smallskip
        {\color{utilitygreen}\textbf{Utility.} Unnecessary support costs 100;
        failing to intervene before a dropout costs 1,000, so the implied
        threshold is \(p^\star=100/(100+1000)=0.091\).\par}
      \end{minipage}
      \hfill
      \begin{minipage}[t]{0.48\linewidth}
        \textbf{Condition A: false-positive cost withheld}\par\smallskip
        {\color{domainblue}\textbf{Domain/covariates.} The same case.\par}\smallskip
        {\color{framingorange}\textbf{Scenario.} The event of interest is
        whether the student drops out. Choose \textsc{Intervene} or
        \textsc{Hold}. \emph{The intervention takes about 45 minutes to carry
        out.} If the event occurs, \textsc{Hold} leaves you 1,000 worse off
        than \textsc{Intervene} because the preventable loss was missed.\par}\smallskip
        {\color{utilitygreen}\textbf{Required output.} Nothing states what an
        unnecessary intervention costs, so no threshold exists. The correct
        answer is A: report a probability, flag the missing false-positive
        cost, and make no decision.\par}
      \end{minipage}
    \end{minipage}%
  }
  \endgroup
  \caption{One abridged case as a complete-information item (left) and under
  condition A of the decidability task (right). Colors identify the components
  of each prompt. The implied threshold \(p^\star\) is shown for the reader and
  never appears in a prompt; italics mark the length-matched irrelevant
  sentence that replaces the withheld cost, so the two prompts differ only in
  whether that cost is recoverable.}
  \label{fig:examples}
\end{figure}

\subsection{Complete decision models and generalization}
\label{sec:complete-models}

\paragraph{Prompting conditions.}
We query each instruction-tuned model in three ways. \emph{Direct} requests
only a structured probability and action. \emph{Chain-of-thought} (CoT) asks
the model to estimate the event probability, map the payoffs to the actions,
and compare their expected values. \emph{In-context} prepends to every item the
same framing-neutral worked example, in generic \textsc{Act}/\textsc{Hold}
language with illustrative values, which specifies the procedure without
revealing a test case or framing.

\paragraph{Complete-information SFT.}
For every framing holdout, we train separate Direct and CoT models on a mix of
probability-only and complete-information decision prompts. For case \(x\), the
probability target is the style-matched base-model belief \(p_E(x)\) and the
action target is \(d^\star(p_E)\). Supervision therefore preserves the model's
independently elicited belief and teaches the decision rule it implies; it
never substitutes the dataset label, because the goal is coherence between
belief and action rather than agreement with the label. Direct targets contain only
the final probability and action; CoT targets take the base elicitation trace
whose probability is closest to \(p_E\), set its final probability to
\(p_E\), and append a deterministic calculation with the current framing's
payoffs.

\paragraph{Training, validation, and test sets.}
Five domains are assigned entirely to the test set; within each of the other
15, source items are split 60/20/20 into training, validation, and in-domain
test sets, with repetitions of an item never crossing sets.

We train eight framing-holdout configurations, each holding out two framings,
so that every framing is excluded once under each regime. The four
\emph{full-class} holdouts remove an entire structural class from training:
\{intervention, insurance\}, \{betting, trading\}, \{contingency, holdout\}, or
\{loss-ratio, utility-ratio\}. The four \emph{cross-class} holdouts,
\{betting, utility-ratio\}, \{holdout, intervention\}, \{insurance, trading\},
and \{contingency, loss-ratio\}, keep in training a \emph{structural sibling}
of each held-out framing, that is, another framing from the same class. Each
trained model is tested on 3,200 decision prompts (the in-domain test set and
all prompts from the five unseen domains), so the eight configurations give
25,600 test prompts per SFT style, crossing domain status with framing status.

\subsection{Decision decidability under incomplete information}
\label{sec:decidability}

We next withhold the payoff information needed to recover one or both error
costs. Each case--utility--framing item is assigned to one of four conditions:
A withholds the false-positive cost, B the false-negative cost, C both, and D
neither (the complete-information case). D receives half of the items and
A--C divide the remainder evenly; repetitions share the assignment and the case
evidence is unchanged. Every response reports a probability and a condition
label, and a decision is required only under D. Withholding a cost does not
shorten the prompt: the sentence that would carry it is replaced by one of
matched length about an irrelevant administrative detail of the same framing
(Figure~\ref{fig:examples}, right), so the two versions differ only in whether
the cost is recoverable.

\paragraph{Decision Decidability SFT.}
For each framing holdout, we train Direct and CoT models on the four-condition
task. They use the same belief targets and data splits as the
complete-information models. Targets for A--C identify the missing cost and
omit the action; targets for D include \(d^\star(p_E)\). CoT targets are
assembled from templates: a deterministic sentence stating which costs the
scenario provides and which condition therefore applies, the base model's own
elicitation trace for the belief (as in the complete-information targets), and,
only under D, the same deterministic payoff calculation. The In-context
baseline instead demonstrates all four conditions with one framing-neutral
example.

\paragraph{Routed system.}
To separate recognizing decidability from executing the decision, we compose
two fine-tuned models into a \emph{routed system}. The Direct SFT decidability
model serves as the gate: predictions A--C terminate with its probability and
condition label, and a prediction of D forwards the item, in its
complete-information format, to the split-matched CoT SFT complete-information
model of Section~\ref{sec:complete-models}, which reports its own belief and
action without seeing the gate's. Routing uses the predicted label, so a
misrouted incomplete item stays incomplete, and both terminating a D item and
routing an A--C item count as end-to-end errors.

\subsection{Evaluation measures}
\label{sec:measures}

\paragraph{Complete-information measures.}
We measure the two failure modes separately. \emph{Belief distortion error} is
the absolute change between the reference
belief \(p_E\), elicited from the evidence alone, and the decision-time belief
\(p'\), elicited after actions and payoffs are introduced,
\begin{equation}
  \mathrm{Belief\ Distortion\ Error}
  =\frac{1}{N_1}\sum_{i=1}^{N_1}|p'_i-p_{E,i}|,
  \label{eq:error1}
\end{equation}
over the \(N_1\) items with finite parsed values for both beliefs. Because the
evidence is identical across the two prompts, this measures stability of the
model-relative belief, not calibration. Sampling at temperature \(0.7\) gives
even an undistorted model a positive value, since \(p'\) is one draw and
\(p_E\) the mean of five. Decoding at temperature zero would remove this noise
for our open models, but production models are served with a hidden temperature
the user cannot set, so we instead measure the noise and report the metric
relative to a \emph{sampling-noise floor}. Let \(\sigma_C\) and \(\sigma_A\)
be the sampling spreads of a decision-time belief and of a single elicitation,
both measured by resampling the same prompts (Appendix~\ref{app:noise-floor}).
Under zero distortion, \(p'-p_E\) is approximately Gaussian with variance
\(\sigma_C^{2}+\sigma_A^{2}/5\), and the mean absolute value of a centered
Gaussian is \(\sqrt{2/\pi}\) times its standard deviation, giving
\begin{equation}
  \mathrm{floor}=\sqrt{\tfrac{2}{\pi}}\;\sqrt{\sigma_C^{2}+\sigma_A^{2}/5}.
  \label{eq:noise-floor}
\end{equation}
Figures plot the \emph{excess belief distortion error},
\begin{equation}
  \mathrm{Excess}=\sqrt{\max\!\bigl(\mathrm{Belief\ Distortion\ Error}^{2}-\mathrm{floor}^{2},\,0\bigr)}.
  \label{eq:excess}
\end{equation}
Because distortion and sampling noise are independent, they add as squares, so
the floor is removed in quadrature. The floor is a reference, not a correction:
subtracting it cannot recover the true distortion, and the clamp at zero is a
display choice. The raw metric and floor for every point are in the
supplementary files. \emph{Belief-conditional decision error} is the share
of parsed decisions that disagree with the action optimal under the belief
reported in the same response,
\begin{equation}
  d_{\mathrm{self},i}=d^\star(p'_i),
  \qquad
  \mathrm{Belief\text{-}Conditional\ Decision\ Error}
  =\frac{1}{N_2}\sum_{i=1}^{N_2}
  \mathbb{I}[d_{\mathrm{model},i}\neq d_{\mathrm{self},i}].
  \label{eq:error2}
\end{equation}
Because the reference action is computed from \(p'\), this metric does not
penalize distorted beliefs; it isolates whether the model applies the specified
utilities to the belief it actually reports.

\paragraph{Decidability measures.}
Information accuracy is the fraction of items whose predicted label matches
the assigned condition; joint accuracy additionally requires, on true-D items,
that the decision equal \(d^\star(p')\), and missing required outputs count as
incorrect. Belief-conditional decision error is restricted to true-D items, the
only condition in which a decision is defined, and belief distortion error is
computed on every item that reports a belief, with its own floor measured on
the decidability prompts. For the routed system, \(p'\) is the gate's belief on
terminated routes and the specialist's on routed ones, each compared with the
reference belief of the model that produced it. Confidence intervals are 95\%
percentile intervals from 100 bootstrap resamples.

\section{Results}
\label{sec:results}

\subsection{Setup}
\label{sec:results-setup}

We evaluate six instruction-tuned checkpoints: Gemma~4 E2B, E4B, and 12B
(5.1B, 8B, and 12B total parameters, the counts used in the figures) and
Qwen3.5 2B, 4B, and 9B. For each we compare three prompting baselines on the base model
(Base Direct, Base CoT, and Base In-context), the two SFT styles, and the
routed system of Section~\ref{sec:decidability}. SFT results pool the eight framing holdouts (25,600 test prompts per model)
and are split by framing status (seen, held-out cross-class, held-out
full-class) and by domain status (seen, held-out). Base models see every
framing, so their reference values are computed on the same prompts and
reweighted to the domain mix of the SFT test set. Training and decoding details
are in Appendix~\ref{app:training-details}.

\subsection{Complete-information decisions}
\label{sec:results-complete}

Figure~\ref{fig:complete-panels} reports both complete-information measures for
the fine-tuned models against their base models.

\newcommand{\panelrow}[3]{%
  \begin{subfigure}[t]{\linewidth}
    \centering
    \begin{minipage}[c]{0.07\linewidth}
      \centering\normalsize\bfseries(#1)
    \end{minipage}%
    \begin{minipage}[c]{0.93\linewidth}
      \centering\includegraphics[width=\linewidth]{#2}
    \end{minipage}
    \phantomcaption
    \label{#3}
  \end{subfigure}%
}
\begin{figure}[t]
  \centering
  \includegraphics[width=.75\linewidth]{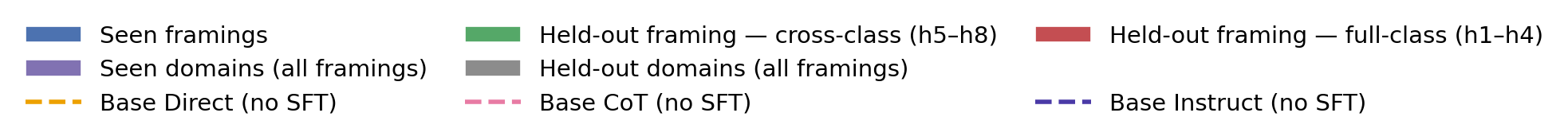}\\[1pt]
  \panelrow{a}{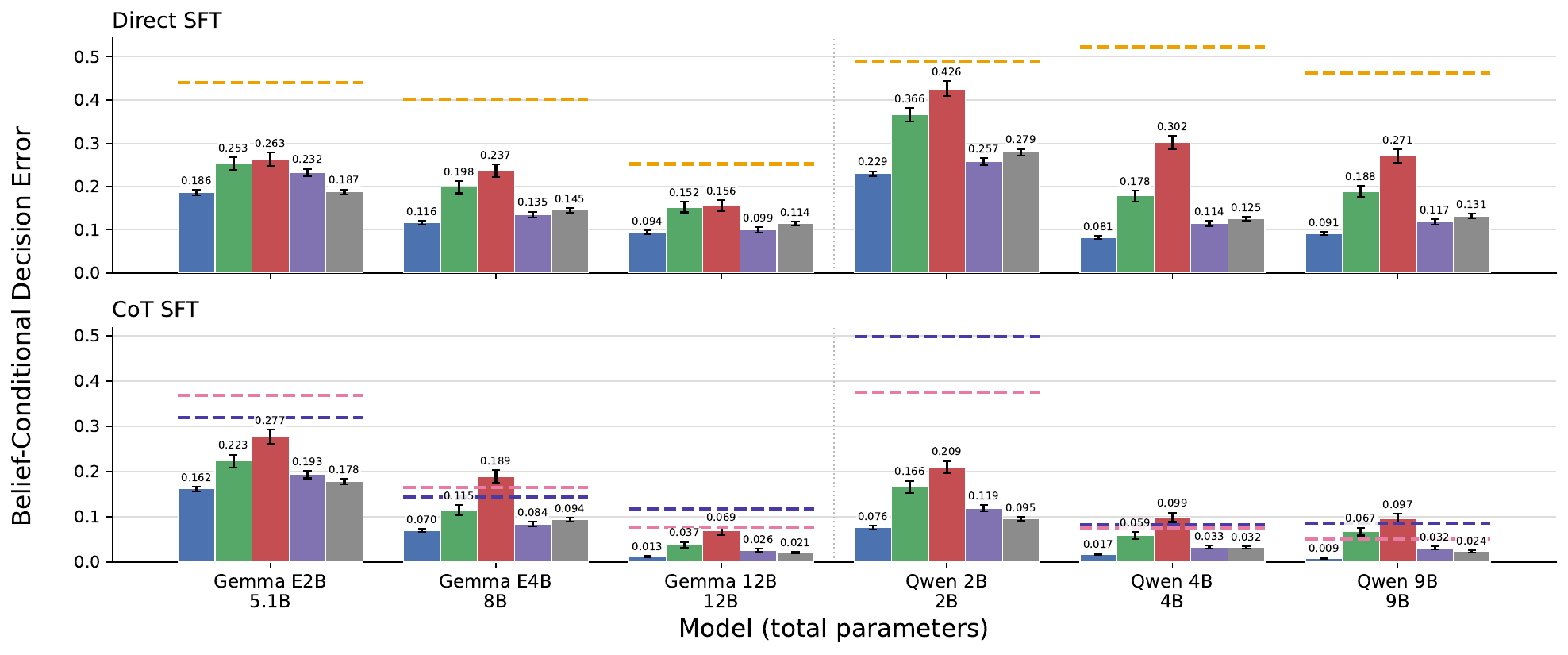}{fig:complete-error2}
  \panelrow{b}{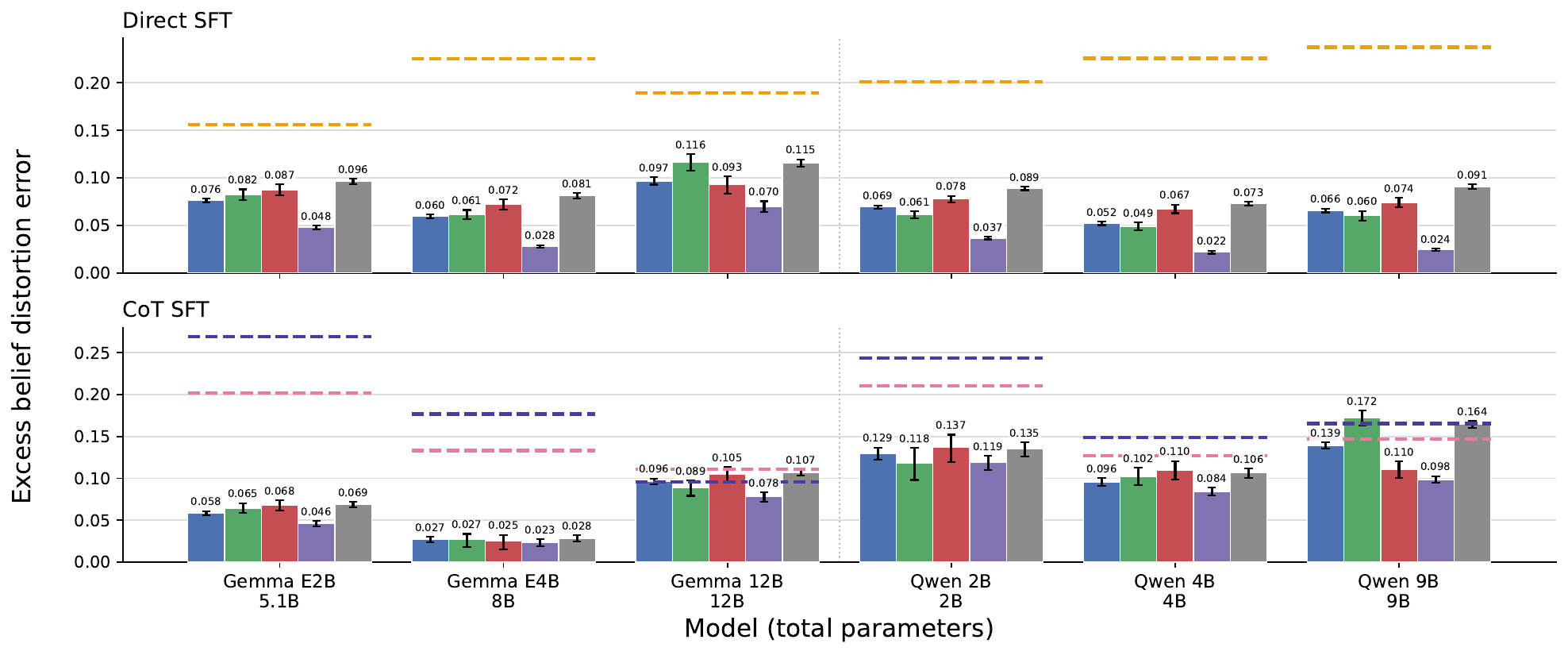}{fig:complete-error1}
  \caption{Complete-information results for Direct SFT (top row of each panel)
  and CoT SFT (bottom row), pooled over the eight framing holdouts: (a)
  belief-conditional decision error (Equation~\ref{eq:error2}); (b) excess
  belief distortion error (Equation~\ref{eq:excess}). The three left bars split
  test prompts by framing status (seen, held-out cross-class, held-out
  full-class) and the two right bars by domain status. Dashed lines are the
  same base model without SFT on the same prompts, reweighted to the test set's
  domain mix. Error bars are 95\% bootstrap intervals.}
  \label{fig:complete-panels}
\end{figure}

\paragraph{Applying the utilities.}
Under Direct prompting, between a quarter and a half of every base model's
decisions contradict the action implied by its own stated belief, and this
error barely falls with scale (Figure~\ref{fig:complete-error2}, top).
\emph{Direct SFT lowers the error for every model on every slice}, to well
below the Base Direct reference. The gain transfers cleanly across domains:
held-out-domain error is within 0.03 of seen-domain error for five of the six
models and below it for the sixth. Transfer across framings is harder: error
rises from seen to cross-class to full-class framings for every model, and
full-class error remains substantial (0.16 to 0.43). A structural sibling in training clearly helps, and even the worst Direct SFT
slice stays below the base reference.

\textbf{CoT SFT yields the most coherent decisions of any system, but its
advantage over prompting shrinks with scale.} For Gemma~12B and the two larger Qwen models, the fine-tuned CoT model
misapplies the utilities on fewer than 2\% of seen-framing items, and CoT SFT
beats Direct SFT on every slice of every model except one. For the smaller
models, Base CoT and Base In-context still err on 15--50\% of items, which CoT
SFT cuts several-fold. For Gemma~12B, Qwen~4B, and Qwen~9B, Base CoT is already
fairly coherent (0.05--0.08); CoT SFT still improves on it for seen framings and
both domain slices, but on held-out full-class framings Gemma~E4B, Qwen~4B, and
Qwen~9B are worse than their own base CoT prompts. Every CoT SFT bar that fails to beat both references is a held-out framing:
recognizing a new payoff structure is what the CoT models learn least well.

\paragraph{Preserving beliefs.}
Introducing payoffs moves the base models' beliefs well beyond sampling noise:
excess belief distortion is 0.16--0.24 for Base Direct and 0.10--0.27 for Base
CoT and Base In-context (Figure~\ref{fig:complete-error1}). \emph{Direct SFT
reduces it to 0.02--0.12 on every slice of every model.} The transfer pattern is the reverse of the decision rule's: held-out framings
raise distortion by at most 0.02, while held-out domains raise it two- to
threefold. Applying the utilities is sensitive to how the problem is framed;
preserving the belief is sensitive to what it is about. CoT SFT lowers
distortion below both references for the four smaller models, but for
Gemma~12B and Qwen~9B base CoT prompting already preserves beliefs about as
well, and several CoT SFT bars sit at or above the reference lines. Direct SFT
preserves beliefs better than CoT SFT for Qwen, and the reverse tends to hold
for Gemma.

\subsection{Decision decidability}
\label{sec:results-decidability}

\begin{figure}[t]
  \centering
  \includegraphics[width=.95\linewidth]{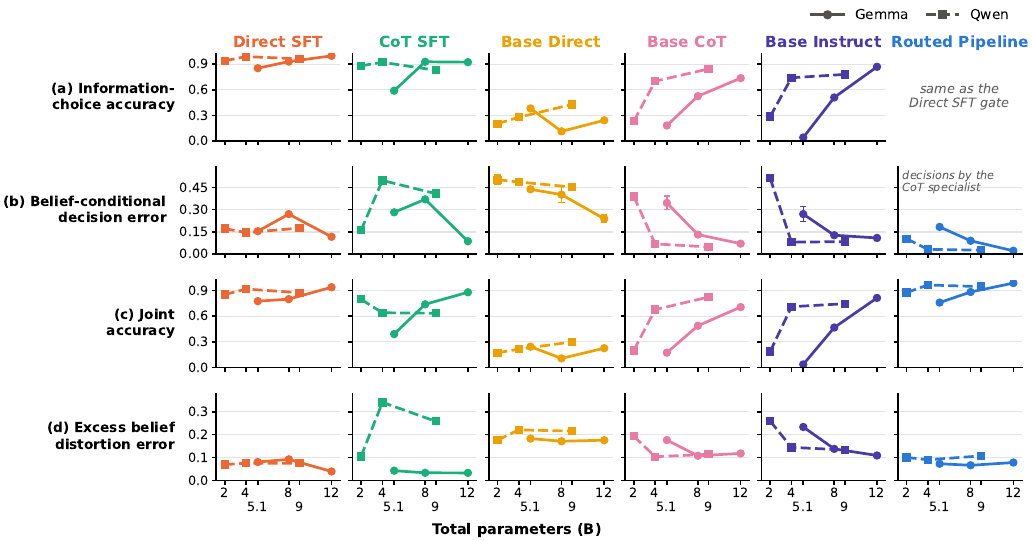}
  \caption{Decidability results by model size (solid: Gemma~4; dashed:
  Qwen3.5). Rows: (a) information-choice accuracy over all four conditions,
  (b) belief-conditional decision error on true-D items, (c) end-to-end joint
  accuracy, and (d) excess belief distortion error over all items that report a
  belief, against a floor measured on the decidability prompts. The routed
  system's information accuracy is its gate's (a), its decisions are the
  complete-information CoT specialist's (b), and its floor is the
  route-weighted mixture of its components' floors (d). Error bars are 95\%
  bootstrap intervals, often smaller than the markers.}
  \label{fig:decidability}
\end{figure}

Figure~\ref{fig:decidability} reports the decidability task, in which half of
the items withhold at least one error cost needed for a coherent decision.

\paragraph{Recognizing missing information.}
Base models rarely notice that a cost has been withheld
(Figure~\ref{fig:decidability}a): Base Direct identifies the correct condition
on only 11--43\% of items, and CoT and In-context prompting help as models grow but never exceed 87\%. \emph{Direct SFT
makes recognition reliable}: 85\% for Gemma~E2B and at least 93\% for every
other model, reaching 99.7\% for Gemma~12B. CoT SFT is never better and often
worse (59--93\%), so the short structured output is the better vehicle for this
skill.

\paragraph{Deciding on complete items.}
On complete items a decidability model must also act on its belief
(Figure~\ref{fig:decidability}b), and here the picture
changes. Direct SFT improves on Base Direct for every model but still misapplies the
utilities on 12--27\% of complete items, more than base CoT prompting does for
all but the two smallest models. CoT SFT trained on the four-condition task is
usually worse still (9--50\%). Asking one model to diagnose the problem and
compute the decision in one response degrades the decision step. The
complete-information CoT specialist, which only ever sees complete items, has
no such difficulty: on the items the gate routes to it, its error is 2--18\%.
\emph{This division of strengths motivates the routed system}: the Direct SFT
gate is the best recognizer of missing information, the complete-information
CoT model the best decision-maker, and routing composes them without additional
training.

\paragraph{End-to-end accuracy.}
Joint accuracy requires the right condition label and, on complete items, the
right decision (Figure~\ref{fig:decidability}c). No base model exceeds 30\% under Direct prompting, and CoT and In-context
prompting reach about 70\% only for the larger models. Direct SFT reaches
78--94\%, with CoT SFT trailing it for every model. \emph{The routed system is
the most accurate for five of the six models} (88--99\%) and within two points
of Direct SFT on Gemma~E2B; since it shares its gate with Direct SFT, the whole
gain comes from handing complete items to the specialist.

\paragraph{Belief stability under incomplete information.}
Fine-tuning again stabilizes beliefs (Figure~\ref{fig:decidability}d): Direct
SFT has low excess distortion for every model (0.04--0.09), and CoT SFT is
lower still for Gemma but unstable for Qwen, where the 4B and 9B models distort
beliefs more than the base prompts do. The routed system lies between its two
components, within 0.04 of Direct SFT for every model, so \emph{it pays little
in belief stability for its gains in decision accuracy}.

\section{Discussion}
\label{sec:discussion}

A language model that estimates risk well can still decide badly: across six
small models, base prompting often chose an action that contradicted the
belief reported in the same response and shifted that belief once payoffs
appeared. Supervised fine-tuning on a small factorial benchmark repaired both failures,
and the gains transferred to unseen domains and framings, with Direct SFT
improving every model on every slice and CoT SFT producing the most coherent
decisions wherever its base model was weak. Under incomplete information, a Direct SFT gate learned to recognize a
missing cost reliably, and routing complete cases to a complete-information
specialist gave the highest end-to-end accuracy for five of six models at little
cost in belief stability. These results rest on an evaluation that holds the evidence fixed and varies
only the utilities and their framing, separating belief distortion from utility
misapplication, two failures that final-answer accuracy conflates and that
prior work on calibration, belief coherence, and revealed preferences has
studied only in isolation from training. Measuring distortion against a sampling-noise floor makes the framework
applicable to models whose decoding temperature cannot be controlled, and
length-matched distractors separate recognition of decision relevance from
prompt length. The benchmark covers binary decisions with two error costs on cases rendered
from tabular data, and the reference belief is model-relative, so the measures
certify coherence rather than accuracy. CoT SFT also generalized least well to payoff structures absent
from training, and its advantage over strong base prompting narrowed with
scale. Future work should extend the procedure to richer action and outcome spaces,
teach models to acquire the information they identify as missing, and test
whether the same routing helps larger and closed models.

\subsection*{Reproducibility statement}
The decision rule and reference belief are specified in
Section~\ref{sec:decision-setting} (Equations~\ref{eq:threshold}
and~\ref{eq:reference-belief}), and the two coherence measures with their
sampling-noise floor in Section~\ref{sec:measures}
(Equations~\ref{eq:error1}--\ref{eq:error2}). The SFT targets, data splits,
framing holdouts, decidability conditions, and routed system are described in
Sections~\ref{sec:complete-models} and~\ref{sec:decidability}. Every prompt
template, including the belief elicitation prompt, the eight
complete-information framings, the decidability question, and the
length-matched sentences that replace withheld costs, is given verbatim in
Appendix~\ref{app:prompts}, and the 20 domains with their role sentences,
events, and public dataset identifiers are listed in Appendix~\ref{app:domains}.
Training hyperparameters, decoding settings with the seeding scheme that keeps
repeated samples independent, and the sampling-noise floor calibration are in
Appendix~\ref{app:training-details}. Code implementing the full pipeline (case
construction, SFT data generation, training, inference, scoring, bootstrap
intervals, and figure generation), together with the case lists, model outputs,
and the raw metric and floor for every reported point, will be released with
the paper.

\subsection*{AI use statement}
In this work, we used generative AI tools to help implement and run the
experimental pipeline (writing and refactoring analysis and plotting code) and
for light editing of the text (grammar, spelling, and word choice). We have not
used generative AI tools to generate research ideas or to write the substantive
content of the paper, and the remaining disclosure categories are not applicable
to this work. Large language models are also the objects of study; the model
outputs used as data are fully specified by the prompts in
Appendix~\ref{app:prompts}. We have reviewed all AI-assisted work: AI-generated
code was checked against the original analysis notebooks and its outputs were
verified by the authors. We take responsibility for the final content of this
work, including text, claims or artifacts produced with the aid of generative
AI.

\paragraph{Funding} BW and KY were supported in part by the Army Research Office under Grant Number W911NF-25-1-0281 and the AI Research Institutes Program funded by the National Science Foundation under AI Institute for Societal Decision Making
(AI-SDM), Award No. 2229881. The views and conclusions contained in this document are those of the authors and should not be interpreted as representing the official policies, either expressed or implied, of the Army Research Office or the U.S. Government. The U.S. Government is authorized to reproduce and distribute reprints for Government purposes notwithstanding any copyright notation herein.

\bibliography{iclr2026_conference}
\bibliographystyle{iclr2026_conference}

\appendix
\section*{Appendix}
\section{Training and inference details}
\label{app:training-details}

\paragraph{Models.}
We use six instruction-tuned checkpoints spanning two families and three sizes
each: Gemma~4 E2B, E4B, and 12B, and Qwen3.5 2B, 4B, and 9B. The same data
construction, training protocol, and decoding protocol are applied to every
model; nothing is tuned per model except the sequence and generation budgets
reported below, which are set by the length of each style's targets. Base
weights are always used in bfloat16.

\paragraph{Adapters and quantization.}
SFT is QLoRA. The base model is loaded in 4-bit NF4 with double quantization
and bfloat16 compute dtype, and its weights stay frozen; only the LoRA
parameters are trained. Adapters use rank 16, scaling \(\alpha=32\), dropout
0.05, no bias terms, and a causal language-modeling task type. They are
attached to the attention and feed-forward projections of the language tower
(\texttt{q\_proj}, \texttt{k\_proj}, \texttt{v\_proj}, \texttt{o\_proj},
\texttt{gate\_proj}, \texttt{up\_proj}, \texttt{down\_proj}; for Qwen3.5 the
fused input and output projections of the linear-attention blocks are included
as well). Target modules are matched only inside the language model, so the
multimodal towers of these checkpoints are never adapted.

\paragraph{Optimization.}
Training runs for two epochs with AdamW at a constant learning rate of
\(2\times10^{-4}\) after 15 warmup steps, weight decay 0, and the default
gradient-norm clip of 1.0. The per-device train and eval batch size is 2 with
four gradient-accumulation steps, giving an effective batch of 8 sequences on
a single GPU. Computation is in bfloat16 with gradient checkpointing
(non-reentrant), sequence packing disabled, and seed 124 for the data order,
sampler, and initialization. Every run trains a fresh adapter from the base
checkpoint; no adapter is initialized from another.

\paragraph{Sequence construction and loss.}
Each training row is a two-turn conversation rendered with the model's own chat
template, and the loss is the standard causal language-modeling loss over the
rendered sequence. Rows are never truncated: an example longer than the
sequence cap aborts the run rather than silently dropping the end of a target,
so every supervised target is complete. The cap is 2,048 tokens for the Gemma
models and for the Direct styles of every model, 2,176 tokens for the Gemma~4
E2B decidability CoT models, and 8,192 tokens for the Qwen3.5 CoT models, which
is where the traces are longest.

\paragraph{Validation and checkpointing.}
Each run holds out the split-matched validation file described in
Section~\ref{sec:complete-models} and evaluates on it every 50 optimizer steps;
training loss is logged every 10 steps and checkpoints are written every 200
steps. Runs resume from the latest checkpoint only when the run fingerprint
matches, so a preempted job continues rather than restarting with a different
configuration. The adapter exported at the end of training is the one used for
all reported evaluations.

\paragraph{Number of runs.}
For each base model we train four families of adapters --- complete-information
Direct and CoT, and decidability Direct and CoT --- for each of the eight
framing holdouts, that is 32 adapters per model and 192 adapters in total. The
routed system introduces no additional training: it reuses the decidability
Direct gate and the complete-information CoT specialist from the same holdout.

\paragraph{Hardware and software.}
Training and inference each run on a single 80\,GB A100, one job per adapter or
per evaluation shard, on Azure Machine Learning. The training image pins
Transformers 5.10.0 with TRL, PEFT, Accelerate, and bitsandbytes; inference uses
vLLM 0.23 with the FlashAttention backend, tensor parallel size 1, eager
execution, and GPU memory utilization between 0.90 and 0.92.

\paragraph{Decoding.}
All generation --- base-model belief elicitation, the prompting baselines, and
SFT evaluation --- uses vLLM with temperature \(0.7\), top-\(p=0.95\), and base
seed 124. The seed for an individual request is the base seed combined with a
checksum of the item identifier. Because vLLM is deterministic for a fixed seed
and prompt, a single shared seed would collapse repeated samples of the same
prompt to byte-identical text; deriving the seed from the identifier, which
encodes the repetition index, keeps the five belief elicitations per case and
the two decision repetitions per condition independent while leaving the whole
evaluation reproducible.

\paragraph{Context and generation budgets.}
Budgets are set per style so that no response is cut off mid-answer. Direct
prompts are served with a 4,096-token context and at most 128 new tokens, which
is ample for the structured final block. Gemma CoT and In-context prompts are
served with a 4,096- to 8,192-token context and up to 3,500 new tokens for the
complete-information style and 2,048 for the decidability style. Qwen3.5
prompts are served with a 16,384-token context and up to 8,192 new tokens for
the 4B and 9B models, and 6,000 for the 2B base model. Generation is batched at
32 prompts for the long CoT styles and up to 256 for the Direct styles; each
batch is checkpointed atomically and validated against the expected identifiers
in order, so an interrupted evaluation resumes without reordering or
duplicating rows. Fine-tuned models are served by attaching the adapter to the
base checkpoint through vLLM's LoRA path, one adapter per job at maximum rank
16, so the evaluated weights are exactly the trained ones.

\paragraph{Qwen thinking mode.}
The Qwen3.5 chat template is applied with \texttt{enable\_thinking=false} for
both belief elicitation and evaluation, so the reasoning that we score is the
CoT the prompt or the SFT target asks for rather than an unscored internal
trace. The runner verifies that the rendered prompt actually closes the
thinking span and fails rather than silently generating in reasoning mode.

\paragraph{Sampling-noise floor calibration.}
\label{app:noise-floor}
The spreads in Equation~\ref{eq:noise-floor} are measured, not assumed. For
every model, prompting style, and holdout adapter, we resample 200 test prompts
(75 from seen domains, 125 from held-out domains, matching the test split) a
second time with a different seed but otherwise the same protocol as the
original evaluation, and pair each new decision-time belief with the stored one
by prompt identifier. The mean absolute difference between the two draws of
the same prompt, divided by \(\sqrt{2}\), gives \(\sigma_C\) up to the Gaussian
constant; the same construction on the five belief-elicitation samples gives
\(\sigma_A\). Base prompting styles use the two decision repetitions per
condition that the base evaluation already contains. Confidence intervals on
each floor come from a hierarchical bootstrap of 10,000 replicates that
resamples holdout adapters and then prompts within adapter, keeping the two
draws of a prompt paired. Floors are distribution-free in the reported pairwise
mean absolute difference and rely on the Gaussian conversion only for the
\(\sqrt{2/\pi}\) constant; direct-style beliefs fall on a coarse grid with
about half of all pairs identical, so this conversion is approximate, and the
raw metric accompanies every excess value in the supplementary files. The
decidability figures use
a separate floor measured on the decidability prompts, whose belief
distribution differs from the complete-information one. Held-out and seen
domains receive their own floors on the domain-split figure.

\section{Prompts}
\label{app:prompts}

Every item is a single user turn, rendered with the model's own chat template.
The templates below are reproduced with their placeholders in braces:
\texttt{setup} is the domain's one-sentence role framing
(Appendix~\ref{app:domains}), \texttt{covariate\_prose} is the case evidence
written as prose, \texttt{event\_E\_description} is the event clause, and
\texttt{instruction} is the response-style instruction. The same evidence and
event strings appear verbatim in the elicitation prompt and in all eight
framings of the same case, so nothing but the payoff presentation changes
across framings. For display, long lines are wrapped and the em dashes of the
source are shown as double hyphens; the strings are otherwise exact.

\subsection{Response-style instructions}

Three instructions are substituted into the instruction placeholder. Direct
forbids any text outside the final block; CoT invites reasoning first; In-context
prepends a framing-neutral worked example and then asks for the same
step-by-step reasoning. The CoT instruction differs between belief elicitation,
where no payoffs exist, and the decision prompts, where the model is asked to
carry its probability into an expected-value comparison.
\begin{promptbox}{CoT, belief elicitation}
\begin{verbatim}
First, think step by step about the relevant evidence and factors. Then give
your final answer wrapped in a single <final> block, in exactly this format:
\end{verbatim}
\end{promptbox}

\begin{promptbox}{Direct, all prompts}
\begin{verbatim}
Output ONLY the final answer block below and NOTHING ELSE. Do NOT write any
reasoning, analysis, explanation, calculation, working, or preamble -- no text
before or after the block. Your entire response must be exactly this block, in
this format:
\end{verbatim}
\end{promptbox}

\begin{promptbox}{CoT, complete-information decision}
\begin{verbatim}
First, think step by step about the relevant evidence to determine your
probability that the event occurs. Then weigh your options: using your
probability and the costs and payoffs in the case, work out the expected value
of each available choice -- its outcome if the event occurs and its outcome if
it does not -- and pick the choice with the better expected value. Give your
final answer wrapped in a single <final> block, in exactly this format:
\end{verbatim}
\end{promptbox}

\begin{promptbox}{In-context, prefix prepended to the CoT decision instruction}
\begin{verbatim}
General guidance on how to turn a probability into a decision using the costs
and payoffs given in the case:
  - Compare the available choices by their expected value, treating any cost
  you pay as a negative amount and any payment or gain you receive as a
  positive amount, and choose the option with the higher expected value.
  - For each choice, take its outcome if the event occurs and its outcome if
  it does not, weight each by your probability that the event occurs, and add
  them. A choice with a fixed, certain amount simply has that amount as its
  expected value -- negative if it is a cost you pay, positive if it is a
  payment you receive.
Worked example 1 (illustrative numbers, not this case): suppose your
probability that the event occurs is 0.20, and you must choose between an
active option that costs 50 for certain and a passive option that loses 300
only if the event occurs (and 0 otherwise).
  - Active option: a fixed cost of 50, so its expected value is -50.
  - Passive option: it loses 300 only when the event occurs, so its expected
  value is 0.20 x (-300) + 0.80 x 0 = -60.
  - Comparing -50 (active) with -60 (passive), the active option has the
  higher expected value, so you would choose the active option.
Worked example 2 (illustrative numbers, not this case, where the certain
option is a payment you RECEIVE rather than a cost): suppose your probability
that the event occurs is 0.70, and you must choose between a sure option that
pays you 100 for certain and a contingent option that pays 200 if the event
occurs and 50 if it does not.
  - Sure option: a fixed payment of 100 you receive, so its expected value is
  +100.
  - Contingent option: 0.70 x 200 + 0.30 x 50 = 140 + 15 = 155.
  - Comparing +100 (sure) with +155 (contingent), the contingent option has
  the higher expected value, so you would choose the contingent option.
Apply this same reasoning to the case above using its own numbers.
\end{verbatim}
\end{promptbox}

\subsection{Belief elicitation}

The elicitation prompt contains the evidence and the event, and no actions or
payoffs. It is issued in its own context, five times per case, and the five
parsed probabilities are averaged into the reference belief \(p_E\) of
Equation~\eqref{eq:reference-belief}. The In-context condition elicits with the
plain CoT instruction, because its worked example concerns decisions and there
is no decision to make here.
\begin{promptbox}{Belief elicitation template}
\begin{verbatim}
{setup}

Case context:
{covariate_prose}

Question: What is the probability that {event_E_description}?

{instruction}
<final>
Probability: <a single number between 0 and 1>
</final>
\end{verbatim}
\end{promptbox}

\subsection{Complete-information decision framings}

Each framing is instantiated from a single threshold \(p^\star\) and a base
scale \(G=100\). The certain quantity is always \(G\) (the intervention cost,
premium, stake, contract price, and sure amount); the event-contingent quantity
is \(G/p^\star\) (the averted loss, loss from inaction, insurance payout, and
contingent reward); and the net speculative gain is \(G(1-p^\star)/p^\star\)
(the betting win and contract profit). The two abstract framings carry no
amounts at all: they state the multiplier \(\max(R,1/R)\) with
\(R=p^\star/(1-p^\star)\), phrased so that the costlier error is named first.
Numbers are rounded for readability, and we recompute \(p^\star\) from the
rounded numbers of every rendered prompt as a construction check.
\begin{promptbox}{Framing 1, intervention (protective action)}
\begin{verbatim}
{setup}

Case context:
{covariate_prose}

An intervention costs {cost_intervene}. If you intervene and
{event_E_description} would have occurred, you avert a loss of {averted_loss}.
If it would not have occurred, the intervention cost is wasted. If you do not
intervene and the event occurs, you incur a loss of {loss_inaction}.

{instruction}
<final>
Probability: <a single number between 0 and 1>
Decision: <INTERVENE or HOLD>
</final>
\end{verbatim}
\end{promptbox}

\begin{promptbox}{Framing 2, insurance (protective action)}
\begin{verbatim}
{setup}

Case context:
{covariate_prose}

You can buy protection against {event_E_description} for a premium of
{premium}. If the event occurs, the protection pays out {payout}. If it does
not occur, the premium is lost. If you go uninsured and the event occurs, you
bear the full loss of {payout}.

{instruction}
<final>
Probability: <a single number between 0 and 1>
Decision: <BUY PROTECTION or GO UNINSURED>
</final>
\end{verbatim}
\end{promptbox}

\begin{promptbox}{Framing 3, betting (hedge or offset)}
\begin{verbatim}
{setup}

Case context:
{covariate_prose}

A bookmaker offers a bet on whether {event_E_description}. If you BET YES: you
stake {stake}; you win {win_amount} (net) if the event occurs and lose your
stake if it does not. If you DECLINE, you neither win nor lose.

{instruction}
<final>
Probability: <a single number between 0 and 1>
Decision: <BET YES or DECLINE>
</final>
\end{verbatim}
\end{promptbox}

\begin{promptbox}{Framing 4, trading (hedge or offset)}
\begin{verbatim}
{setup}

Case context:
{covariate_prose}

A trading desk offers a contract on whether {event_E_description}. You can BUY
the contract for a price of {contract_price}: it pays out {contract_win} (net
profit) if the event occurs, and expires worthless (you lose what you paid) if
it does not. If you PASS, you neither gain nor lose.

{instruction}
<final>
Probability: <a single number between 0 and 1>
Decision: <BUY or PASS>
</final>
\end{verbatim}
\end{promptbox}

\begin{promptbox}{Framing 5, contingency (sure vs.\ contingent payoff)}
\begin{verbatim}
{setup}

Case context:
{covariate_prose}

You can take a FLAT FEE of {sure_amount} to handle this case for certain.
Alternatively, work on CONTINGENCY: you are paid {reward_amount} only if
{event_E_description}, and nothing if it does not happen. The flat fee is
yours regardless of the outcome.

{instruction}
<final>
Probability: <a single number between 0 and 1>
Decision: <CONTINGENCY or FLAT FEE>
</final>
\end{verbatim}
\end{promptbox}

\begin{promptbox}{Framing 6, holdout (sure vs.\ contingent payoff)}
\begin{verbatim}
{setup}

Case context:
{covariate_prose}

A buyer offers a firm {sure_amount} for the asset if you CASH OUT today.
Alternatively, you can HOLD OUT for a payment of {reward_amount} that is
realized only if {event_E_description}, and nothing if it does not. Cashing
out pays {sure_amount} for certain.

{instruction}
<final>
Probability: <a single number between 0 and 1>
Decision: <HOLD OUT or CASH OUT>
</final>
\end{verbatim}
\end{promptbox}

\begin{promptbox}{Framing 7, loss-ratio (abstract ratio)}
\begin{verbatim}
{setup}

Case context:
{covariate_prose}

This is a binary decision between ACT and HOLD for {event_E_description}. No
specific amounts are given; only the relative cost of the two possible
mistakes matters: {loss_ratio_clause} A correct decision costs nothing. Choose
the option with the lower expected cost.

{instruction}
<final>
Probability: <a single number between 0 and 1>
Decision: <ACT or HOLD>
</final>
\end{verbatim}
\end{promptbox}

\begin{promptbox}{Framing 8, utility-ratio (abstract ratio)}
\begin{verbatim}
{setup}

Case context:
{covariate_prose}

This is a binary decision between ACT and HOLD for {event_E_description}. Each
correct decision earns you a positive reward, and only the relative size of
these rewards matters: {utility_ratio_clause} A wrong decision earns nothing.
Choose the option with the higher expected reward.

{instruction}
<final>
Probability: <a single number between 0 and 1>
Decision: <ACT or HOLD>
</final>
\end{verbatim}
\end{promptbox}

\subsection{Decision decidability prompts}

The decidability prompts reuse the same evidence, event, and framings, but
replace the framing paragraph with a scenario built from three sentences and
add the four-way information question. The first sentence names the event and
the two actions. The second describes what the false-positive error costs and
the third what the false-negative error costs, each stated as how much worse
off one action leaves you than the other in that state of the world. Under
conditions A, B, or C the corresponding sentence is replaced by a sentence of
matched length about an irrelevant administrative detail of the same framing,
so the withheld and present versions are equally long and equally specific and
differ only in whether the cost is recoverable.
\begin{figure}[t]
  \centering
  \begingroup
  \setlength{\fboxsep}{7pt}
  \fcolorbox{black!25}{black!2}{%
    \begin{minipage}{0.95\linewidth}
      \footnotesize
      \begin{minipage}[t]{0.475\linewidth}
        \textbf{Condition A: false-positive cost withheld}\par\smallskip
        {\color{domainblue}\textbf{Domain/covariates.} A 27-year-old student
        entered with a rank of 120/200, does not receive a scholarship, and
        passed 0/4 first-semester courses and 0/5 second-semester
        courses.\par}\smallskip
        {\color{framingorange}\textbf{Scenario.} The event of interest is
        whether the student drops out. Choose \textsc{Intervene} or
        \textsc{Hold}. \emph{The intervention takes about 45 minutes to carry
        out.} If the event occurs, \textsc{Hold} leaves you 1,000 worse off
        than \textsc{Intervene} because the preventable loss was
        missed.\par}\smallskip
        {\color{utilitygreen}\textbf{Required output.} The miss cost is given,
        but nothing states what an unnecessary intervention costs, so the
        false-positive cost is unrecoverable and no threshold exists. The
        correct answer is A: report a probability, flag the missing
        false-positive cost, and make no decision.\par}
      \end{minipage}
      \hfill
      \begin{minipage}[t]{0.475\linewidth}
        \textbf{Condition B: false-negative cost withheld}\par\smallskip
        {\color{domainblue}\textbf{Domain/covariates.} A machine is operating
        at 301.2 K air temperature and 310.3 K process temperature, at
        1,379 rpm and 43.2 Nm torque, with 56 minutes of tool wear.\par}\smallskip
        {\color{framingorange}\textbf{Scenario.} The event of interest is
        whether the machine fails. Choose \textsc{Buy Protection} or
        \textsc{Go Uninsured}. If the event does not occur,
        \textsc{Buy Protection} leaves you 100 worse off than
        \textsc{Go Uninsured} because the premium was unnecessary.
        \emph{Going uninsured adds 4 forms to the case file.}\par}\smallskip
        {\color{utilitygreen}\textbf{Required output.} The premium is given,
        but nothing states what an uninsured failure costs, so the false-negative
        cost is unrecoverable and no threshold exists. The correct answer is B:
        report a probability, flag the missing false-negative cost, and make no
        decision.\par}
      \end{minipage}
    \end{minipage}%
  }
  \endgroup
  \caption{Two abridged incomplete items. Italics mark the length-matched irrelevant
  sentence that replaces the withheld cost. In each case one error cost is
  stated and the other is absent, so \(p^\star\) cannot be formed and no action
  is justified at any belief.}
  \label{fig:decidability-examples}
\end{figure}

\begin{promptbox}{Decidability template}
\begin{verbatim}
{setup}

Case context:
{covariate_prose}

{scenario}

Question: What is the probability that {event_E_description}, and is there
enough information to choose between the options?

A) Not enough information about false positives (acting when the event does
not occur).
B) Not enough information about false negatives (not acting when the event
occurs).
C) Not enough information about either false positives or false negatives.
D) Enough information is given. Choose D and provide the better decision.

{instruction}
<final>
Probability: <a single number between 0 and 1>
Choice: <A, B, C, or D>
Decision: <{act_word} or {hold_word} if Choice D; N/A otherwise>
</final>
\end{verbatim}
\end{promptbox}

The action words in the last line are the framing's own, so the response
format never reveals which framing class is in play beyond what the scenario
already says. The style instructions are decision-specific: CoT is asked to
state the applicable choice before reasoning about the evidence, and In-context
receives a worked example that walks through all four conditions on one generic
case.
\begin{promptbox}{CoT, decidability}
\begin{verbatim}
First inspect the decision information. Begin your reasoning by stating which
Choice A, B, C, or D applies and explain why: identify whether the
false-positive cost, the false-negative cost, both, or neither are missing.
Then reason about the case evidence and state your probability that the event
occurs. For Choice D only, compare the options using that probability and
provide the better decision. Give your final answer wrapped in a single
<final> block, in exactly this format:
\end{verbatim}
\end{promptbox}

\begin{promptbox}{Direct, decision decidability}
\begin{verbatim}
Output ONLY the final answer block below and NOTHING ELSE. Do NOT write any
reasoning, analysis, explanation, calculation, working, or preamble. Select A,
B, or C when the specified information is missing. Select D only when both
kinds of information are available, and then provide the better decision. Your
entire response must be exactly this block, in this format:
\end{verbatim}
\end{promptbox}

\begin{promptbox}{In-context, prefix prepended to decision decidability instruction}
\begin{verbatim}
General worked example (the same example is used for every framing):
Suppose a generic case asks you to choose ACT or HOLD for an event, and the
case evidence leads you to assign probability 0.30 to the event.
  1. Initially, no information is given about what is lost by ACT when the
  event does not occur (the false-positive or false-alarm cost), and no
  information is given about what is lost by HOLD when the event occurs (the
  false-negative or miss cost). Both are missing, so the answer is Choice C.
  2. Now add that a false negative costs 80, while still giving no
  false-positive cost. Information about false positives is missing, so the
  answer is Choice A.
  3. Return to the original both-missing case. Instead add that a false
  positive costs 20, while giving no false-negative cost. Information about
  false negatives is missing, so the answer is Choice B.
  4. Finally, give both costs: a false positive costs 20 and a false negative
  costs 80. There is enough information, so use Choice D and make the
  decision. At probability 0.30, ACT has expected error cost (1 - 0.30) x 20 =
  14, while HOLD has expected error cost 0.30 x 80 = 24. ACT has the lower
  expected error cost, so the completed answer is Choice D with Decision ACT.

\end{verbatim}
\end{promptbox}

The scenario sentences for all eight framings are listed below, instantiated
at \(p^\star=1/6\) with base scale \(G=100\), which gives a false-positive cost
of 100 and a false-negative cost of 500. Condition D uses both ``present''
sentences, A replaces the false-positive sentence, B the false-negative
sentence, and C both.
\begin{promptbox}{Scenario sentences by framing}
\begin{verbatim}
[intervention]  Decision field: <INTERVENE or HOLD>
  opening    : The event of interest is whether {event_E_description}. Choose
  INTERVENE or HOLD.
  FP present : If the event does not occur, INTERVENE leaves you 100 worse off
  than HOLD because the intervention was unnecessary.
  FP withheld: The intervention takes about 45 minutes to carry out.
  FN present : If the event occurs, HOLD leaves you 500 worse off than
  INTERVENE because the preventable loss was missed.
  FN withheld: A review of held cases is logged within 30 days.

[insurance]  Decision field: <BUY PROTECTION or GO UNINSURED>
  opening    : The event of interest is whether {event_E_description}. Choose
  BUY PROTECTION or GO UNINSURED.
  FP present : If the event does not occur, BUY PROTECTION leaves you 100
  worse off than GO UNINSURED because the premium was unnecessary.
  FP withheld: The protection policy runs for a term of 12 months.
  FN present : If the event occurs, GO UNINSURED leaves you 500 worse off than
  BUY PROTECTION because the uncovered loss was missed.
  FN withheld: Going uninsured adds 4 forms to the case file.

[betting]  Decision field: <BET YES or DECLINE>
  opening    : A bookmaker offers a bet on whether {event_E_description}.
  Choose BET YES or DECLINE.
  FP present : If the event does not occur, BET YES leaves you 100 worse off
  than DECLINE because the stake is lost.
  FP withheld: BET YES must be placed at least 6 hours before the event is
  resolved.
  FN present : If the event occurs, DECLINE leaves you 500 worse off than BET
  YES because the net win was missed.
  FN withheld: DECLINE keeps an account record on file for 3 years.

[trading]  Decision field: <BUY or PASS>
  opening    : A trading desk offers a contract on whether
  {event_E_description}. Choose BUY or PASS.
  FP present : If the event does not occur, BUY leaves you 100 worse off than
  PASS because the contract price is lost.
  FP withheld: The offset contract settles 5 days after expiry.
  FN present : If the event occurs, PASS leaves you 500 worse off than BUY
  because the net gain was missed.
  FN withheld: PASS produces a brokerage note that is 8 pages long.

[contingency]  Decision field: <CONTINGENCY or FLAT FEE>
  opening    : The event of interest is whether {event_E_description}. Choose
  CONTINGENCY or FLAT FEE.
  FP present : If the event does not occur, CONTINGENCY leaves you 100 worse
  off than FLAT FEE.
  FP withheld: Taking the FLAT FEE requires 3 hours of paperwork.
  FN present : If the event occurs, FLAT FEE leaves you 500 worse off than
  CONTINGENCY.
  FN withheld: CONTINGENCY work is reviewed every 6 weeks.

[holdout]  Decision field: <HOLD OUT or CASH OUT>
  opening    : The event of interest is whether {event_E_description}. Choose
  HOLD OUT or CASH OUT.
  FP present : If the event does not occur, HOLD OUT leaves you 100 worse off
  than CASH OUT.
  FP withheld: CASH OUT must be completed within 4 business days.
  FN present : If the event occurs, CASH OUT leaves you 500 worse off than
  HOLD OUT.
  FN withheld: HOLD OUT requires 2 signatures to maintain.

[lossratio]  Decision field: <ACT or HOLD>
  opening    : This is a binary decision between ACT and HOLD for whether
  {event_E_description}. A correct decision costs nothing; choose the option
  with lower expected cost.
  FP present : If the event does not occur, ACT leaves you 1 normalized units
  worse off than HOLD.
  FP withheld: The assessment form contains 10 administrative fields.
  FN present : If the event occurs, HOLD leaves you 5 normalized units worse
  off than ACT.
  FN withheld: The decision record is retained for 7 years.

[utilityratio]  Decision field: <ACT or HOLD>
  opening    : This is a binary decision between ACT and HOLD for whether
  {event_E_description}. A wrong decision earns nothing; choose the option
  with higher expected reward.
  FP present : If the event does not occur, ACT leaves you 1 normalized reward
  units worse off than HOLD.
  FP withheld: The reward schedule is reviewed every 9 months.
  FN present : If the event occurs, HOLD leaves you 5 normalized reward units
  worse off than ACT.
  FN withheld: The decision record requires 5 approval signatures.
\end{verbatim}
\end{promptbox}

\section{Domains}
\label{app:domains}

Table~\ref{tab:domains} lists 20 explored domains. Each is backed by a public
tabular dataset with a binary outcome; a case renders selected covariates
of one row as prose and defines the event from the dataset outcome. The role
sentence shown for each domain is the literal \texttt{setup} string that opens
every prompt for that domain, and the event clause is the literal string
substituted into every elicitation and decision prompt. The 15 training domains
supply the training, validation, and in-domain test splits; the five held-out
domains appear only at test time and are never seen during SFT, so they measure
transfer of the decision rule to unseen subject matter rather than to unseen
framings.
\begin{table}[H]
  \centering
  \footnotesize
  \begin{tabular}{@{}p{0.145\linewidth}p{0.40\linewidth}p{0.36\linewidth}@{}}
    \toprule
    Domain & Role sentence opening each prompt & Event \(E\) \\
    \midrule
    \multicolumn{3}{@{}l}{\textit{Training domains (15): train, validation, and in-domain test splits}}\\[0.2em]
    Credit-card default risk & You are a credit-risk analyst at a bank reviewing a customer's credit-card account to judge repayment risk. & the cardholder defaults on their next payment \\[0.25em]
    Loan credit risk & You are a loan officer at a retail bank reviewing a new loan application. & the applicant turns out to be a bad credit risk \\[0.25em]
    Corporate bankruptcy & You are a corporate credit analyst assessing whether a company is financially distressed from its accounting ratios. & the company goes bankrupt within the forecast horizon \\[0.25em]
    Term-deposit marketing & You are a marketing analyst at a retail bank judging whether a client will subscribe to a term deposit if called. & the client subscribes to a term deposit if contacted \\[0.25em]
    Income-support targeting & You are an analyst screening census records to identify lower-income individuals for a support program. & the person earns at most \$50K a year (low income) \\[0.25em]
    Student dropout & You are an academic advisor reviewing a university student's record to gauge their risk of dropping out. & the student drops out before completing their degree \\[0.25em]
    Hospital readmission & You are a hospital discharge planner reviewing a diabetic patient's stay to gauge their readmission risk. & the patient is readmitted within 30 days of discharge \\[0.25em]
    Breast-tumour diagnosis & You are a pathologist reviewing cell-nucleus measurements from a breast fine-needle aspirate. & the breast tumour is malignant \\[0.25em]
    Predictive maintenance & You are a reliability engineer monitoring an industrial machine's live sensor readings. & the machine fails during operation \\[0.25em]
    Online article popularity & You are a content strategist at an online news publisher forecasting whether a newly published article will be widely shared. & the article becomes widely shared (popular) \\[0.25em]
    Steel-plate fault detection & You are a quality-control inspector judging, from automated surface measurements, whether a flagged region of a steel plate is physically scratched. & the inspected region is a genuine surface scratch in the steel (not a clean or otherwise-marked surface) \\[0.25em]
    Email spam filtering & You are an email-security analyst judging, from word and character statistics, whether a message was sent as unsolicited bulk email. & this email was sent as unsolicited bulk email (spam) rather than by a legitimate correspondent \\[0.25em]
    Phishing-website detection & You are a web-security analyst judging, from a site's technical signals, whether it was created to deceive users and steal their credentials. & this website was built to steal users' credentials (a phishing site) rather than being a legitimate site \\[0.25em]
    Telecom customer churn & You are a customer-retention analyst at a telecom company reviewing a subscriber's account. & the customer churns next cycle \\[0.25em]
    Employee attrition & You are an HR analyst reviewing an employee's profile to gauge their risk of leaving the company. & the employee leaves the company within the year \\[0.25em]
    \midrule
    \multicolumn{3}{@{}l}{\textit{Held-out domains (5): test only, never seen during SFT}}\\[0.2em]
    Wildfire occurrence & You are a wildfire analyst assessing fire risk from current weather conditions in a region of Algeria. & a fire occurs under these conditions \\[0.25em]
    Gamma-ray signal detection & You are an astrophysicist classifying a shower image recorded by a ground-based gamma-ray telescope. & the recorded shower is a gamma-ray signal \\[0.25em]
    Room-occupancy detection & You are a smart-building controller deciding whether an office room is currently occupied from its sensor readings. & the room is occupied \\[0.25em]
    Online purchase intent & You are an e-commerce analyst judging whether a website visitor will complete a purchase during their session. & the visitor completes a purchase this session \\[0.25em]
    Heart-disease screening & You are a cardiologist reviewing a patient's diagnostic test results. & the patient has heart disease \\[0.25em]
    \bottomrule
  \end{tabular}
  \caption{Decision domains, role sentences, and events.
  Ten cases are sampled per domain, giving 200 evidence contexts.}
  \label{tab:domains}
\end{table}
Eighteen domains come from the UCI Machine Learning Repository (dataset ids
350, 144, 365, 222, 2, 697, 296, 17, 601, 332, 198, 94, 327, 547, 159, 357, 468, 45); telecom churn and employee attrition come from the public IBM sample CSVs
of the same names. Outcome labels are used to construct and audit cases and are
never used as probability targets for SFT.

\end{document}